# Global Public Health Surveillance using Media Reports: Redesigning GPHIN

Dave CARTER [a], Marta STOJANOVIC [a], Philip HACHEY [a], Kevin FOURNIER [a], Simon RODIER [a], Yunli WANG [a], and Berry DE BRUIJN [a]

[a] *Digital Technologies Research Centre, National Research Council Canada, Ottawa*

**Abstract.** Global public health surveillance relies on reporting structures and transmission of trustworthy health reports. But in practice, these processes may not always be fast enough, or are hindered by procedural, technical, or political barriers. GPHIN, the Global Public Health Intelligence Network, was designed in the late 1990s to scour mainstream news for health events, as that travels faster and more freely. This paper outlines the next generation of GPHIN, which went live in 2017, and reports on design decisions underpinning its new functions and innovations.

## 1. Introduction

Global public health surveillance relies on reporting structures and transmission of trustworthy health reports. But in practice, these processes may not always be fast enough, or are hindered by procedural, technical, or political barriers. At the same time, the news travels much faster and more freely, certainly in this day and age. Recognizing that, Public Health Agency of Canada has been operating the Global Public Health Intelligence Network (GPHIN) since 1997, designed around capturing mainstream media stories from around the world, analyzing them, and then disseminating the insights globally, through WHO, for rapid responses such as initiating treatment and vaccination efforts, and slowing spread through travel advisories and port-of-entry screening.

This paper describes the efforts and results from the GPHIN rejuvenation project.[1] It provides an overview of the (added) functionalities, design decisions, and high-level implementation details. It also aims to inspire readers that are looking to extend these ideas to other types of real-time, broad scope surveillance efforts in life sciences.

## 2. Background

For a detailed overview of disease surveillance principles and systems, we refer to review papers on this topic. Abat et al. in their 2016 review paper [1] put multiple syndromic surveillance efforts in perspective, citing multiple data sources from the environment (in-cluding weather and animal data), population behaviours (including drug sales and internet use), health care (clinics and hospitals), and demographics. We note that mainstream media was not considered in this review. Barboza et al. [2] analyzed several earlier biosurveillance systems and concluded that they are indeed generally effective at publishing information about infectious diseases using news media compared to a gold standard.

---

[1] GPHIN renewal was a team effort. The authors wish to acknowledge the contributions of research, development, operational, and support staff at National Research Council and Public Health Agency of Canada.

In addition to human disease surveillance, there are monitoring systems for livestock health that can inspire us. Lower privacy concerns surrounding animals have contributed to quicker advancements of such systems. Additionally, problems with livestock health can lead to human health concerns downstream, as animals can function as human disease vectors, and livestock epidemics can affect the human food supply chain [9]. EpidVis [7] is one such initiative that combines news collection with compelling visualizations for animal epidemiology.

GeoSentinel is a syndromic surveillance network that collects and disseminates reports on sicknesses in travellers. The information can inform pre-travel preparation, as well as guide post-travel differential diagnosis when a recent traveller has fallen ill [8].

The first version of GPHIN was conceived in the late 1990s, and soon proved its worth by enabling early detection of the 2003 SARS outbreak [6]. While the premise of the system remained relevant, advancements in software technology and data processing methods warranted a ground-up rejuvenation effort, which this paper reports on.

## 3. Methods and Results

The base architecture of GPHIN involves data intake, data processing, information storage, and interactive user tools [3]. In many cases, third-party or off-the-shelf components are used. Data originates from multiple sources, the majority of articles arriving through a large media aggregation service, others through curated RSS feeds – news sites, public health organizations, product recalls, and some high-relevance epidemiological (micro)blogs. Processed and tagged data are then stored in a relational database and also in an off-the-shelf full-text search tool, for analysts to search, retrieve, and digest.

A number of innovations were implemented. We highlight several of these in the following paragraphs, starting with its multilingual capabilities, including translation.

Multiple languages: GPHIN is a global effort, and as it uses local-level mainstream media signals, it must be able to process multiple languages. GPHIN processes data in ten languages: English, French, Spanish, Portuguese, Russian, Arabic, Farsi, Simplified and Traditional Chinese, and Indonesian – covering logograph, abjad and alphabet scripts and covering both left-to-right and right-to-left languages. Text analytics takes place partly in the language of origin (keyphrase extraction from restricted phrase lists), partly based on English machine-translations of the source document. Pragmatically, the Microsoft Azure Translate API was selected to detect the language of source material and translate it to English. These translations are somewhat noisy, and are never used as full substitutes to the source text, but even lossy translation can inform human analysts as well as various machine analytics components such as classifiers. Occasionally, downstream problems such as ambiguity can be averted by using modern context-sensitive machine translation algorithms: for example, the Chinese character can mean *twenty-first*, the family name *Ma*, or *horse*, and only one meaning is relevant when tracking equine diseases.

Relevance scoring is a function that discriminates first-line reports from background stories, retrospectives, opinions, policy documents, or any story that was accidentally caught by the aggregator service (e.g., "Bieber fever"). The technique is implemented as a binary classifier but with a confidence estimate that allows for ranking and flexible threshold setting. The classifier [5] is built around a machine learning classifier (SVM) using a set of approximately 1400 documents represented as unrestricted word and character n-grams. Articles with a high relevance score are published immediately, while low-scoring articles are automatically suppressed; only the remaining medium-relevance articles are triaged by analysts.

Information extraction: Medical terms in text are detected and mapped to UMLS by MetaMap Lite. The tool is restricted to a set of semantic types that is generally relevant to the public health community – largely syndromes and disease vectors. Subsequently, a hand-curated blacklist of common false positive terms is applied. The resulting annotations inform downstream components such as clinical reasoning, and links to UMLS definitions and ontology elements are provided next to articles. General named entities, such as person and organization names, are tagged using Stanford CoreNLP's named entity recognizer. Heuristics were created to infer people's job titles when mentioned, as well as to consolidate organizations' full names and acronyms.

Geographic entity resolution: One part of situational awareness is knowing precisely where things are happening. We use Stanford CoreNLP's named entity recognition plus heuristics to resolve mentions of geographic entities (countries, states/provinces, cities/towns, some lakes, some regions) to latitude-longitude coordinates and to canonical forms in the Geonames ontology so that they can be counted and mapped, collecting synonymous mentions of places and inferring parent-child relationships. When large outlets report on Tokyo or New York, local news sites may specify the sub-city (Yokohama or Shinjuku; Queens or Tribeca). A search in GPHIN for Tokyo will include all sub-cities, which makes it easier to synthesize very local media from a global perspective. This function also reconciles synonyms and language differences, coalescing multiple spellings/forms into a single known entity. Non-unique place names (e.g. London, UK versus London, Ontario, Canada) are resolved through heuristics that take into account where an article was published. Geo-tagging allows users to quickly drill into details about the geography of a document, checking population sizes, median age, poverty rates, and human development index measurements, for example.

Deduplication: Near-duplicate reports on the same incident are unavoidable when many media outlets republish newswire articles verbatim. To avoid duplicate stories hijacking trend detection algorithms, a duplicate detection algorithm was devised. The technique is built on two main principles: text edit distance, where a certain percentage of the word triplets in the article must be the same; and shingling, an observation that if all possible three- or four-word phrases are randomly assigned a number, sorting and sampling the set of such unique numbers for the phrases in two documents produces a substantially similar sample for documents that are similar but not necessarily identical. This solves the challenges like editors lightly rewording newswire articles for space constraints and adding/removing extra paragraphs of context at the end of an article. Duplicates are not removed but clustered / flagged as such, and the initial version among candidate duplicates is used as the exemplar. Minor but critical differences, such as an updated death count, are extracted using StructPred and tracked separately.

Narrative discovery and change point detection: On a daily basis, it looks for statistically unusual pairs of non-generic ontology keywords plus locations, and publishes these in a list of new "narratives". As new articles come in to the system that match each such pair of keyword plus location, it aggregates the content and tracks the narrative. While GPHIN may ingest as many as 10 000 new articles in a given day, the number of new narratives tends to be between 20 to 50 daily, which is small enough for an analyst to digest. For narratives that span at least several days, GPHIN applies topic modelling technology [10] to estimate when the language of the narrative is changing – when the language of an infectious disease pandemic is changing from uncertain to certain, or from investigation to response to rehabilitation, and so on. While GPHIN was designed around data streams that carry more noise than permissible for highly sensitive aberration detection algorithms, this combination of narrative discovery plus change point detection accomplishes a similar goal of trend detection and tracking.

User interface: The dynamic and faceted nature of the task is supported by a range of user interface modes. Figure 1 illustrates some of these. The basic interaction model in GPHIN is a set of filtering criteria coordinated with a colour-coded map, a configurable bar graph (showing activity over time or top keywords/locations/categories), and a list of documents. If a user has a hypothesis to test, a user might compose a search for a tagged syndrome in a geographic area, and then drill down by clicking on red hot spots on the interactive map or drill down into increasingly specific keywords or combinations thereof (Figure 1, left). If the event is a serious outbreak or a mass gathering, there are integrated report generation tools to support daily surveillance workflows outside the tool – users can tag and collate documents into topical reports, and GPHIN analysts can highlight portions of text within documents. Analysts can disseminate their findings via e-mail and RSS, while recipients can specify their own alert filters.

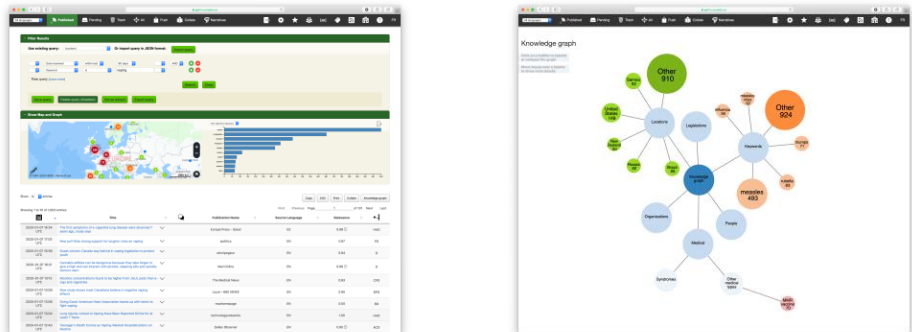

**Figure 1.** Coordinated search, map, and graph tool (left); knowledge graph for a measles outbreak (right)

GPHIN also supports workflows where the user has no specific hypothesis. The recently discovered narratives are listed with recent activity time series graphs, extractive summaries [11], and any detected inflection points. A further tool that GPHIN provides to try to illustrate surprising correlations is a basic knowledge graph (Figure 1, right), placing top keywords, locations, people and medical entities in bubbles. Location bubbles are linked when they are geographically adjacent. GPHIN is used primarily for monitoring current events in recent news, but also supports retrospective research, whether for psychoactive substances [12] or for analysing how media reporting changed the behaviour of the general public in the midst of past pandemics [4].

## 4. Discussion

GPHIN currently has more than one thousand registered user accounts in 56 countries, representing public health, governments at various levels, military, research, education, diplomatic, and travel organizations, from global to municipal. Aside from day-to-day syndromic surveillance activities, GPHIN was used for specific mass gatherings and summits (Olympics, G7), as these tend to come with exceptional volumes and directions of human (and therefore disease) travel. Surrounding the 2016 summer Olympic Games in Rio de Janeiro, WHO epidemiologists were not just concerned about Zika, but also that an ongoing MERS outbreak in the Middle East might migrate to Rio.

As domain knowledge was kept separate from the software code, we have since expanded the platform to support food safety surveillance (food, animal, and plant hazards, for example) and natural disasters in the context of public safety.

## 5. Conclusion

We presented an overview of added functionality to a global syndromic surveillance tool that leverages local mainstream media articles, in a range of different languages, to provide a rapid and global view to public health officials.